\documentclass[twoside,leqno,twocolumn]{article}
\usepackage{ltexpprt}

\usepackage{times}
\usepackage{helvet}
\usepackage{courier}
\usepackage{amsmath}
\usepackage{amssymb}
\usepackage{times}
\usepackage{algorithm}
\usepackage{algorithmicx}
\usepackage{algpseudocode}
\usepackage{multicol, blindtext}
\usepackage{graphicx}  
\usepackage{subcaption}
\usepackage{color}
\usepackage{breqn}

\begin{document}

\title{\Large Learning Solving Procedure for Artificial Neural Network} 
\author{ 
   \large \textbf{Ju-Hong Lee}\\[-5pt]
   \normalsize Department of Computer Engineering  \\[-5pt]
   \normalsize Inha University \\[-5pt]
   \normalsize	juhong@inha.ac.kr \\[-5pt]
  \and
   \large \textbf{Moon-Ju Kang} \\[-5pt]
   \normalsize Department of Computer Engineering \\[-5pt]
   \normalsize Inha University \\[-5pt]
   \normalsize sktel1020@nate.com \\[-5pt]
  \and
   \large \textbf{Bumghi Choi} \\[-5pt]
   \normalsize Department of Computer Engineering \\[-5pt]
   \normalsize Inha University \\[-5pt]
   \normalsize bgchoi666@gmail.com \\[-5pt]
}
\date{}

\maketitle







\begin{abstract} \small\baselineskip=9pt 
It is expected that progress toward true artificial intelligence will be achieved through the emergence of a system that integrates representation learning and complex reasoning (LeCun et al. 2015). In response to this prediction, research has been conducted on implementing the symbolic reasoning of a von Neumann computer in an artificial neural network (Graves et al. 2016; Graves et al. 2014; Reed et al. 2015). However, these studies have many limitations in realizing neural-symbolic integration (Jaeger. 2016). Here, we present a new learning paradigm: a learning solving procedure (LSP) that learns the procedure for solving complex problems. This is not accomplished merely by learning input-output data, but by learning algorithms through a solving procedure that obtains the output as a sequence of tasks for a given input problem. The LSP neural network system not only learns simple problems of addition and multiplication, but also the algorithms of complicated problems, such as complex arithmetic expression, sorting, and Hanoi Tower. To realize this, the LSP neural network structure consists of a deep neural network and long short-term memory, which are recursively combined. Through experimentation, we demonstrate the efficiency and scalability of LSP and its validity as a mechanism of complex reasoning.
\end{abstract}

\section{Introduction}
As an attempt to implement human reasoning capabilities with a machine, an expert system with a knowledge base and an inference engine has emerged. However, this has proven insufficient for transplanting human complex reasoning into a machine, which, as a top-down approach, has reached its limit. A new approach called ''connectionism,'' which uses an artificial neural network (ANN), has emerged. ANN, using the back-propagation algorithm of a multi-layer architecture (Werbos. 1974), has been applied to many recognition areas. However, notable achievements have been made via algorithm improvements (Hahnloser, et al. 2000; Srivastava, et al. 2014; Hochreiter, et al. 1997) and parallel computing, using multiple CPUs and GPUs that enable the use of many layers, or steps, in feed-forward or recurrent neural networks. Thus, problems of overfitting (Tetko, et al. 1995) and vanishing gradient (Hochreiter. 1991; Hochreiter, et al. 2001) can be resolved. Remarkable achievements have also been realized in the areas of image recognition (Krizhevsky, et al. 2012; Farabet, et al. 2013; Tompson, et al. 2014; Szegedy, et al. 2014), speech recognition (Mikolov, et al. 2011; Hinton, et al. 2012; Sainath, et al. 2013), and natural language processing (Bordes, et al. 2014).
However, most of these achievements represent learning by input-output data-processing, which falls short of the intent to implement complex reasoning. Von Neumann computers, of course, possess a solid history as dependable symbolic-reasoning machines. However, they simply and faithfully execute prewritten reasoning algorithms, rather than demonstrate a complex ability to reason creatively like humans. Therefore, classical artificial intelligence, such as expert systems, expose the weakness in autonomous learning due to the lack of adaptable capability.

Alternatively, advanced ANNs, such as deep learning networks, demonstrate excellent learning capabilities via input-output patterns. However, ANNs experience difficulty learning complex computational or logical reasoning problems that are otherwise easily handled by von Neumann computers. Therefore, several attempts have been made to implement the symbolic reasoning of a von Neumann computer with an ANN architecture, leveraging the idea that neural-symbolic integration (Jaeger. 2016) will be needed for complex reasoning. A modeling study of biological neuron performance in working memory using continuous firing to solve explicit tasks (Hazy. 2006; Dayan. 2008; Eliasmith. 2013) has been conducted in the field of neuroscience. A. Graves introduced the neural Turing machine (Graves et al. 2014), which implements these concepts with ANNs. The neural Turing machine allows read-write operations in a given external memory space, enabling weighted addressing through long short-term memory (LSTM) (Hochreiter et al. 1997), as with a Turing machine or a general computer. Its extension is a differentiable neural computer (DNC) (Graves, et al. 2016), which means external memory is treated as a variable when differentiable read-write controls are activated. This is similar to associative long-term potentiation of the mammalian hippocampus (Graves, et al. 2016). DNC has the effect of creating adaptability by transplanting, into an ANN, the symbolic reasoning mechanism of a von Neumann computer, (i.e., Turing machine) in which processes and memory are separated. However, the experiment, consisting of synthetic questions and answers, graphs, and block puzzles, showed that a DNC does not have the ability to solve complex reasoning problems like human beings (Jaeger. 2016). Learning through a DNC is not about learning a sequence of procedures or sub-problems (i.e., tasks), but is instead about learning a set of data transitions and transformations (e.g., memory usage and temporal linkage) for a given complex problem.

Learning by input-output data relying on the generalization and regularization of a learning method does not constitute a logical reasoning process. Learning by input-output data may not precisely solve a problem, because it does not rule out the possibility of accidently solving with a different algorithm. Take, for example, an instance where the input-output pattern is coincidentally identical for a given sample dataset. Assume that there is an algorithm, $A$, for problem, $P$.
$$A:D\rightarrow R$$

The training set, $S$, is sampled from the input space, $D$, of algorithm,$A:(S\subset D)$. If an ANN, $N$, is trained using training sample, $S$, can we be sure that “$N$ has learned $A$?” The answer is generally, “no.” Generalization of learning is often described as a complement to the completeness of algorithmic learning. With input-output data-learning, why is complete learning of the algorithm impossible? Generally, there exists another problem, $Q$, with algorithm, $B$, that satisfies the following condition for $A$ of $P$.

$$\exists B\:such\:that\: B:D\rightarrow R$$
$$\forall x\in S,A(x)=B(x)$$
$$\exists y\in (D-S)\: such\: that\: A(y)\not =B(y)$$

The output of $A$ and $B$ for sample, $S$, coincide. However, the output of $A$ and $B$ may be different for some of the remaining parts, except for $S$. If so, then which algorithm did $N$ learn, $A$ or $B$? It may not be revealed until testing. It may be neither $A$ nor $B$. Therefore, if the size of input space is extremely large, we can only approximately learn the algorithm of a given problem via input-output; but, it is impossible to learn completely.

In this regard, research in which high-level programs are learned as a series of low-level programs has been conducted with neural programmer-interpreters (NPI) (Reed, et al. 2015). With NPI, 21 programs (i.e., program embedding), environments (i.e., external memory), input-output parameters, etc., are used to train given programs as a series of subprograms. This has the advantage of learning with a smaller amount of sample data than with the input-output pattern of DNC. In terms of implementing ANN for the entire program logic, rather than just a neural simulation of the external memory system (i.e., DNC), we make the program the object of learning. Add and Sort functions can be learned as a decomposed series of low-level sub-functions. It is somewhat inadequate that the programs to be studied are limited, and the execution of each sub-program required in the learning process is independent of the neural system. Owing to the lack of iterative and recursive expansion methods, there are limits to expanding algorithmic learning to more complex problems.

Here, we propose a new learning paradigm: a learning solving procedure (LSP) that learns a problem-solving capability, the main factor of complex reasoning. LSP provides a way to learn complex algorithms as sequential combinations of simpler problems. Additionally, unlike NPI, LSP uses a symbolic representation freely expressed by numbers and symbols, rather than by rigid form-of-function, which is a learning theme. LSP is composed of repetitively and recursively usable components. It learns and executes the algorithm of a complex problem by combining a deep neural network (DNN) and LSTM. Experiments demonstrate that solving algorithms, such as complex arithmetic expressions of addition and multiplication, sorting, and Hanoi Tower, can be successfully learned and solved with LSP.

\section{Learning Solving Procedure}
\noindent The goal of this study is implementing an ANN that can learn algorithms for given problems and execute the algorithms directly. For algorithm learning, we propose a learning solving procedure (LSP), which is a new learning paradigm that learns the procedure for solving a given problem, and design an architecture that implements LSP. The final LSP architecture can learn and execute well-known complex algorithms.

For a neural network to learn a problem-solving procedure, we first answer the following questions.

\begin{itemize}
\item \textbf{Question 1.} How can the problem-solving procedure be expressed?
\item \textbf{Question 2.} If the neural network learns the algorithm, what structure should it use so that the algorithm can be executed correctly?
\end{itemize}

To answer \textbf{Question 1}, we propose a formal method called ''task.'' There are two types of tasks: simple and complex. A simple task has an immediate answer, and a complex task consists of a series of additional simple or complex tasks. One or more tasks that are required to execute a complex task are called ''subtasks''.

\textbf{Example)} {\fontfamily{qcr}\selectfont
\small{Add(3,4)}} is a simple task with the answer of 7. {\fontfamily{qcr}\selectfont
\small{Add(26,73)}} requires two simple subtasks of {\fontfamily{qcr}\selectfont
\small{Add(6,3)}} for rightmost digit and {\fontfamily{qcr}\selectfont
\small{Add(2,7)}} for the leftmost digit. We discuss ''carrying'' later.

For the neural network architecture of \textbf{Question 2}, we propose an LSP architecture composed of DNN and LSTM. DNN distinguishes whether the input task is simple or complex. For simple tasks, DNN learns from the input-output data and recalls the output for a given input. In the case of a complex task consisting of a series of multiple subtasks, the DNN sends the complex task to the LSTM, which creates a series of subtasks. These subtasks are recursively entered into the LSP to obtain answers. Therefore, the LSP is a pure ANN recursively composed of only DNN and LSTM.

For tasks that can be decomposed into a series of subtasks, LSP can learn the generalized algorithm by learning the procedure to solve the problem. This suggests that it can provide a fundamental starting point for complex reasoning in neural networks. To illustrate the working principle of LSP, we demonstrate how to learn the simplest forms of addition and multiplication algorithms.

\subsection{LSP Task for Addition and Multiplication\newline}

\noindent Here, the numbers for addition and multiplication are purely symbolic data, not quantitative. We design a neural network that learns and executes procedures to solve arithmetic addition and multiplication in a way similar to a small child entering elementary school. Those students learn the concept of single-digit addition and multiplication, and they memorize results as tables. For the addition and multiplication of numbers greater than two digits, the memorized tables are used, and the ''carry'' is sent to the next digit on the left.

\vspace{2.5mm}
\textbf{Addition example)}\begin{flushleft}
Execution of {\fontfamily{qcr}\selectfont
\small{Add(738,859)(carry=0)}} is performed in order as subtasks, {\fontfamily{qcr}\selectfont
\small{Add(8,9)(carry=0)}}, {\fontfamily{qcr}\selectfont
\small{Add(3,5)(carry=1)}}, and {\fontfamily{qcr}\selectfont
\small{Add(7,8)(carry=0)}}.
\end{flushleft}

\vspace{2.5mm}
The multiplication of two numbers over two digits is performed sequentially using the memorized multiplication tables as follows.

\vspace{2.5mm}
\textbf{Multiplication example)}\begin{flushleft}
The execution of {\fontfamily{qcr}\selectfont
\small{Mul(567,834)(carry=0)}} is performed in the order as subtasks, {\fontfamily{qcr}\selectfont
\small{Mul(567,4)(carry=0)}}, {\fontfamily{qcr}\selectfont
\small{Mul(567,3)(carry=226)}} and, {\fontfamily{qcr}\selectfont
\small{Mul(567,8)(carry=192)}}. {\fontfamily{qcr}\selectfont
\small{Mul(567,3)(carry=226)}} is performed in order of subtasks, {\fontfamily{qcr}\selectfont
\small{Mul(7,3)(carry=226)}}, {\fontfamily{qcr}\selectfont
\small{Mul(6,3)(carry=24)}}, {\fontfamily{qcr}\selectfont
\small{Mul(5,3)(carry=4)}}. Here, {\fontfamily{qcr}\selectfont
\small{Mul(7,3)(carry=226)}} is converted to {\fontfamily{qcr}\selectfont
\small{Add(226,21)(carry=0)}} using the memorized multiplication tables. {\fontfamily{qcr}\selectfont
\small{Add(226,21)(carry=0)}} is performed as in the addition example described above.
\end{flushleft}

Here, single-digit addition, such as {\fontfamily{qcr}\selectfont \small{Add(8,9)(carry=0)}} and {\fontfamily{qcr}\selectfont \small{Add(8,9)(carry=1)}}, is regarded as a simple task, because the result is a simple value that requires no further computation. {\fontfamily{qcr}\selectfont \small{Add(738,834)(carry=0)}}, {\fontfamily{qcr}\selectfont \small{Mul(7,3)(Carry=226)}}, etc., are considered complex tasks, because they require recursive calculations involving additional subtasks.

\subsection{LSP Architecture for Addition and Multiplication\newline}
\noindent We propose an LSP architecture that can learn and perform addition and multiplication algorithms. By training the LSP neural network with the addition and multiplication solving procedures, the LSP neural network learns the addition and multiplication algorithms.

The LSP architecture is shown in Figure 1. It is a recursive structure composed of DNN and LSTM. All input tasks are input to the DNN, which identifies whether the task is simple or complex. If it is a simple task, it outputs the answer immediately. DNN is trained to memorize addition and multiplication tables. If it is given a complex task, it passes it to LSTM, which outputs subtasks in the order of the solving procedure that is trained and memorized for the complex task. Subtasks are recursively input to the LSP for their execution and to obtain answers.

\begin{figure}[!ht]
\includegraphics[width=0.45\textwidth,height=\textheight, keepaspectratio]{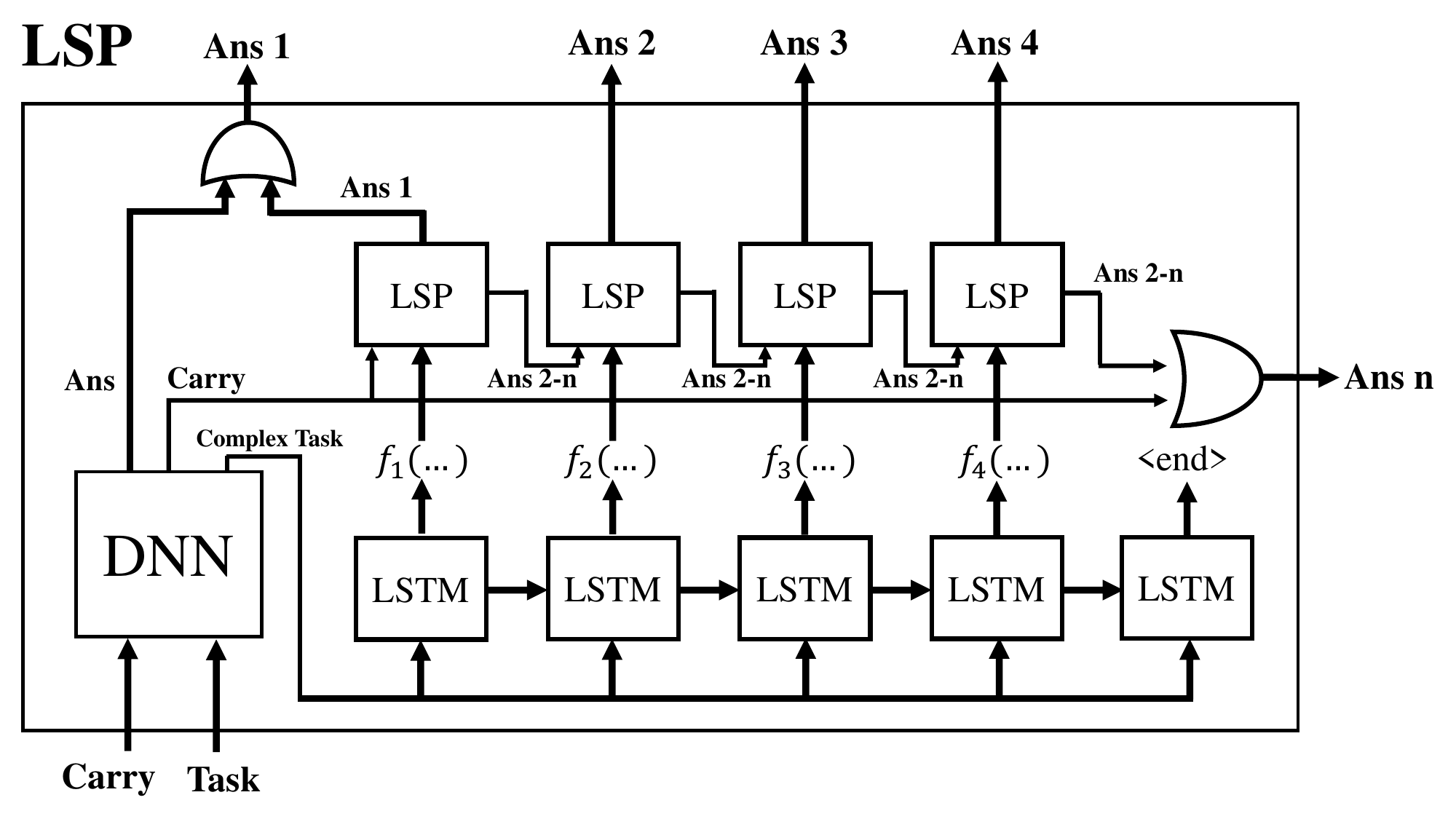}
\caption{LSP Architecture for Addition and Multiplication}
\label{fig:Basic_LSP for Addition and Multiplication}
\end{figure}

\begin{figure}[!ht]
\includegraphics[width=0.45\textwidth,height=\textheight, keepaspectratio]{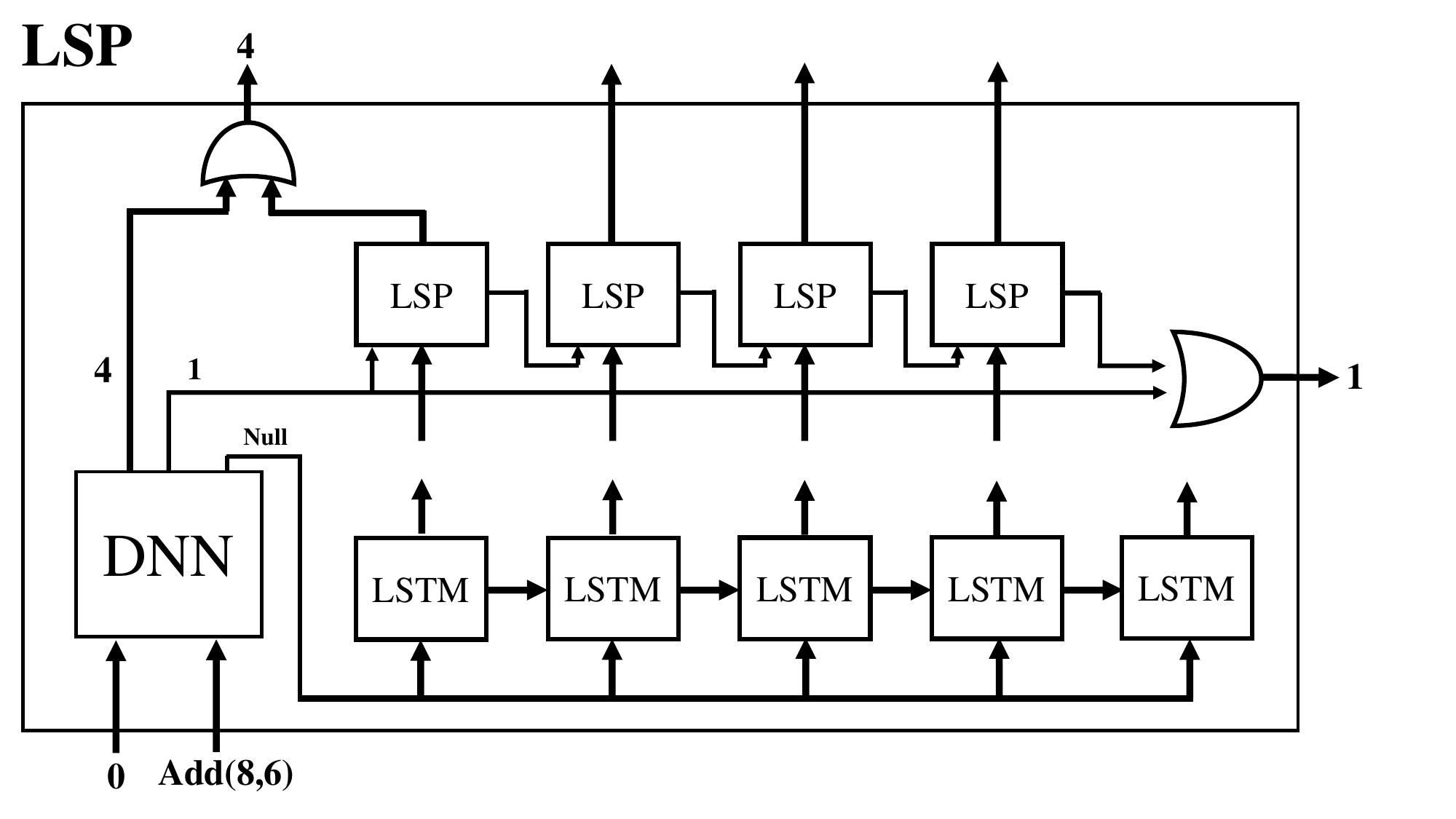}
\caption{An example of a simple task for addition}
\label{fig:Simple Addition}
\end{figure}

\begin{figure}[!ht]
\includegraphics[width=0.45\textwidth,height=\textheight, keepaspectratio]{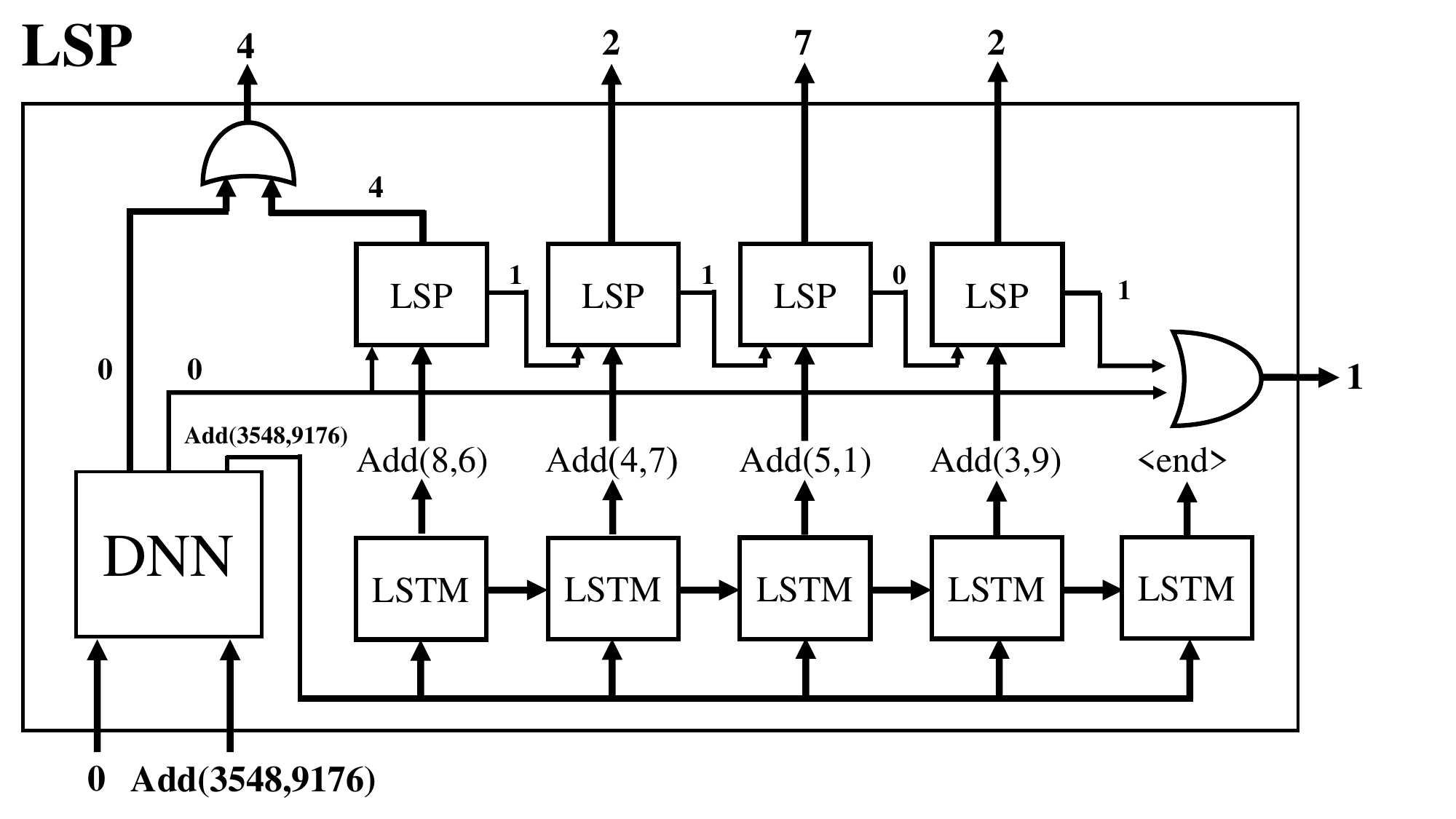}
\caption{An example of a complex task for addition}
\label{fig:Complex Addition}
\end{figure}

\noindent Cases of adding single digits are simple tasks. All others are complex. A complex task, which is the addition of two-digit and larger numbers, is passed to the LSTM, which sequentially generates addition tasks for columnar digits and recursively inputs them into the LSP. The ''carry'' of the LSP is input to consecutive digits to the left.
\vspace{1mm}
\begin{flushleft} 
\par Example: {\fontfamily{qcr}\selectfont
\small{Add(3548, 9176)(carry=0)}} $\rightarrow$ {\fontfamily{qcr}\selectfont
\small{Add(8,6)(carry=0)}}, {\fontfamily{qcr}\selectfont
\small{Add(4,7)(carry=1}}, {\fontfamily{qcr}\selectfont
\small{Add(5,1)(carry=1)}}, and {\fontfamily{qcr}\selectfont
\small{Add(3,9)(carry=0)}}.
\end{flushleft}

\vspace{1mm}
In the case of multiplication, the simple task is a single-digit multiplication without carry. All other cases, including single-digit multiplication with carry, are complex tasks. For complex tasks, LSTM sequentially generates subtasks of the multiplication-solving procedure and processes them recursively in the lower LSP.
\vspace{1mm}

\begin{flushleft}
For example, for {\fontfamily{qcr}\selectfont\small{Mul(2367,4958)(carry=0)}}, the following subtasks are generated as the solving procedure. {\fontfamily{qcr}\selectfont \small{Mul(2367,4958)(carry=0)$\rightarrow$Mul(2367,8)(carry=0), Mul(2367,5)(carry=1893),Mul(2367,9)\\(carry=1372),}} and {\fontfamily{qcr}\selectfont\small{Mul(2367,4)(carry=2267)}}. Again, for {\fontfamily{qcr}\selectfont \small{Mul(2367,5)(carry=1893)}}, the following subtasks are generated as a solving procedure. {\fontfamily{qcr}\selectfont \small{Mul(2367,5)(carry=1893)$\rightarrow$Mul(7,5)(carry=1893), Mul(6,5)(carry=192),Mul(3,5)(carry=22),and Mul(2,5)(carry=3).}} {\fontfamily{qcr}\selectfont \small{Mul(7,5)(carry=1893)}}, which is a one digit multiplication with carry, is transformed into {\fontfamily{qcr}\selectfont \small{Add(1893,0035)(carry=0)}} using multiplication tables memorized by DNN. Then, it is passed to LSTM and processed in the way of addition.
\end{flushleft}

\subsection{Extended LSP Architecture\newline}
\noindent In LSP, for addition and multiplication, we can learn only the solving procedure to compute primitive addition and multiplication problems. We have shown that the LSP model can learn the algorithm (i.e., problem-solving procedure) for specific problems such as addition and multiplication.

However, LSP for addition and multiplication can only perform simple addition and multiplication operations with limited numbers of digits, and is insufficient to learn the task of general problem-solving. Therefore, we extend the LSP architecture for addition and multiplication to design an extended LSP architecture, so that it can solve more complicated problems by adding many LSP-XXXs (i.e., LSP tasks) that effectively compose complex problems and solve them.

\begin{figure}[!ht]
\includegraphics[width=0.45\textwidth,height=\textheight, keepaspectratio]{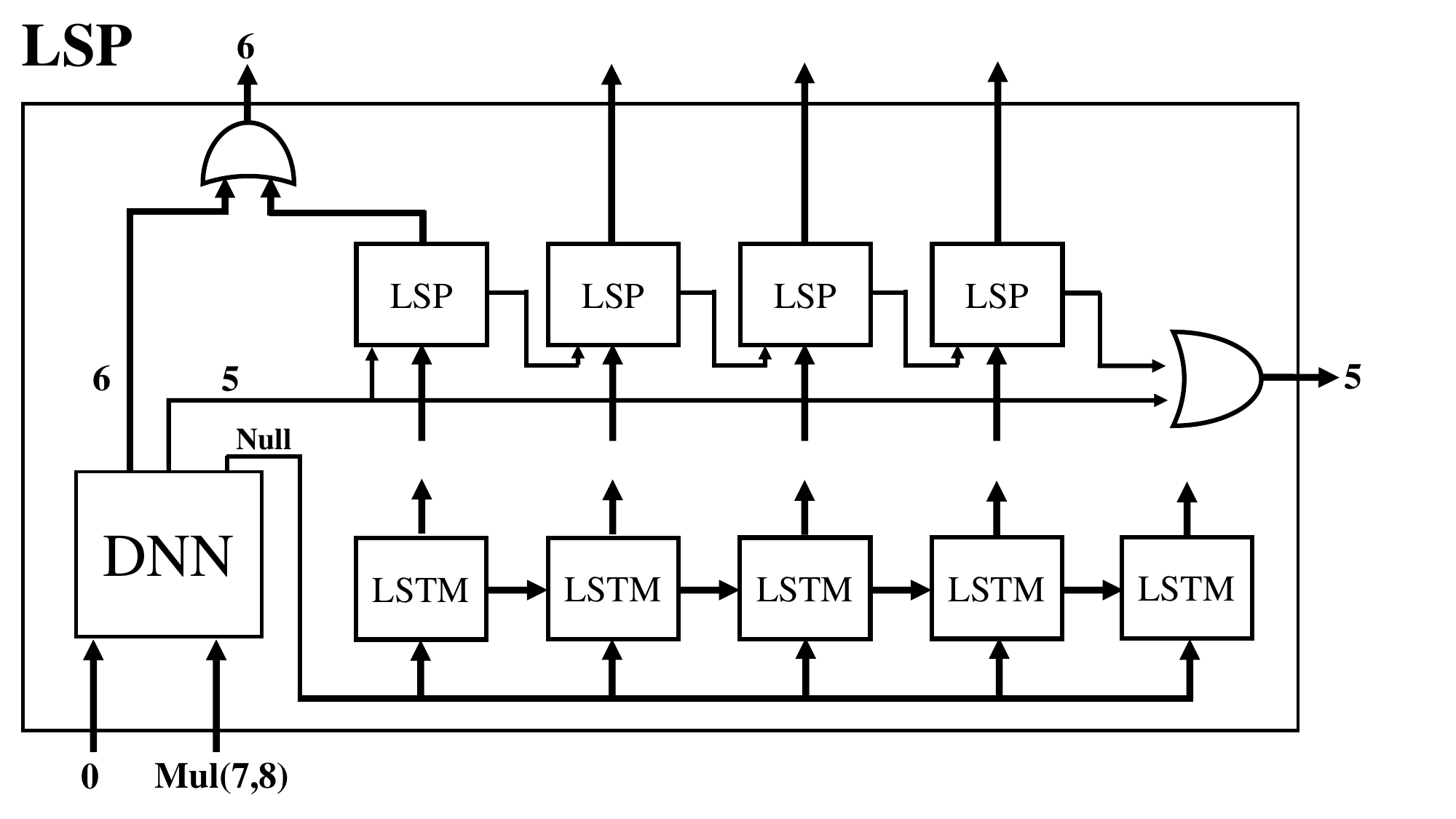}
\caption{An example of a simple task for multiplication}
\label{fig:Simple Multiplication}
\end{figure}

\begin{figure}[!ht]
\begin{subfigure}{.4\textwidth}
    \includegraphics[width=1\textwidth,height=\textheight, keepaspectratio]{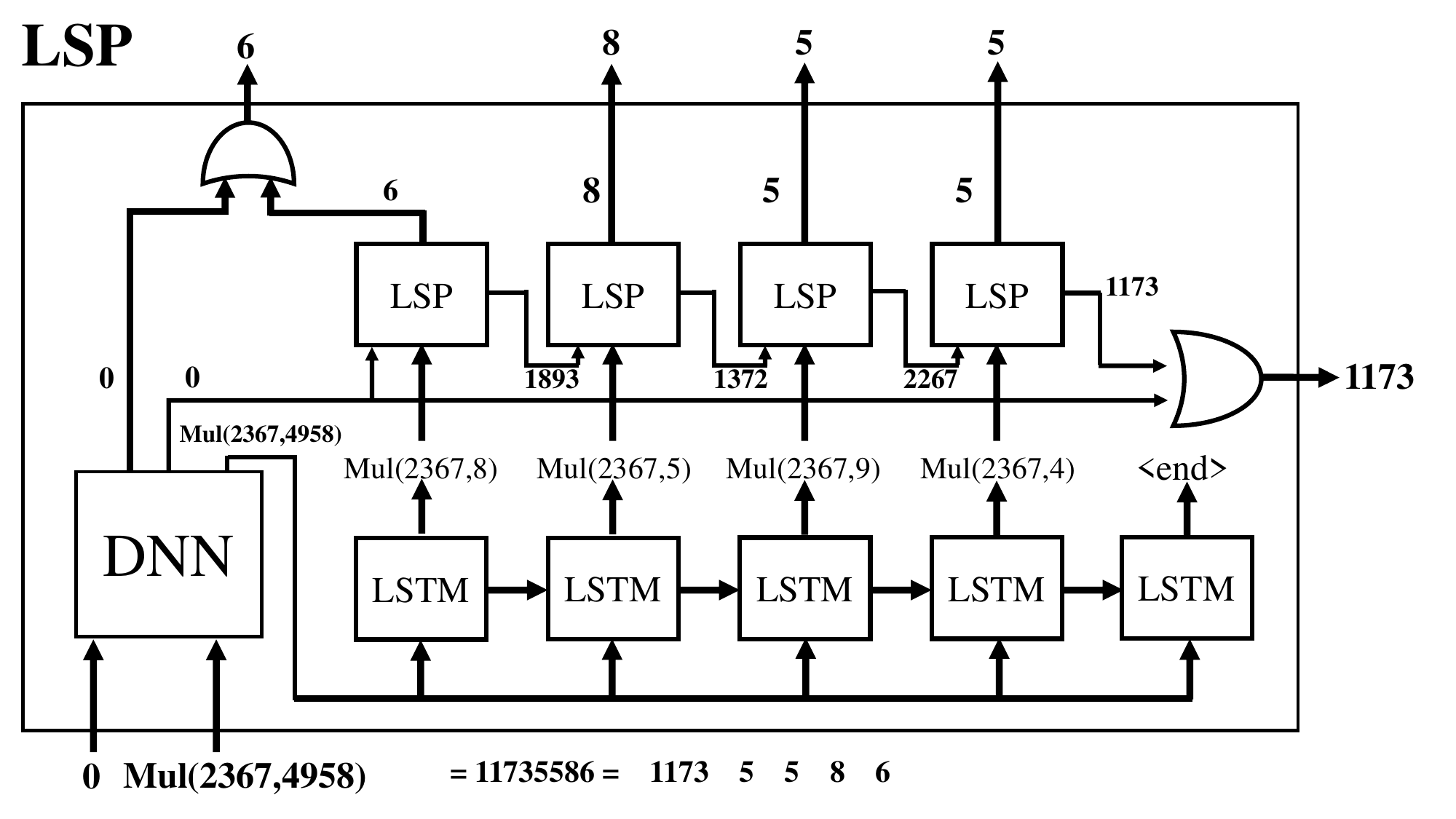}
    \caption{Complex Multiplication 4x4}
    \label{subfig:Complex Multiplication 4x4}
\end{subfigure}
\begin{subfigure}{.4\textwidth}
    \includegraphics[width=1\textwidth,height=\textheight, keepaspectratio]{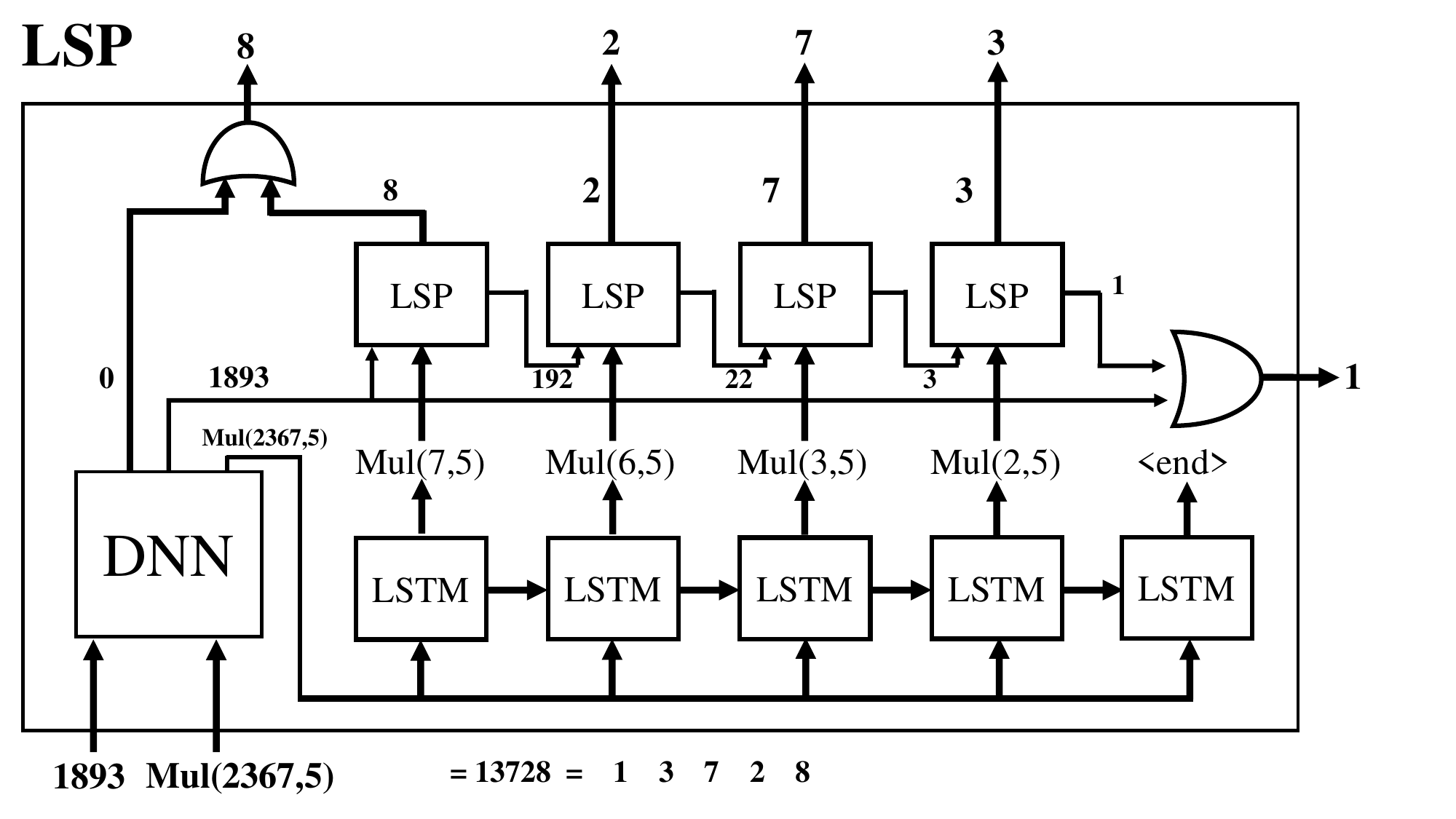}
    \caption{Complex Multiplication 4x1}
    \label{subfig:Complex Multiplication 4x1}
\end{subfigure}
\begin{subfigure}{.4\textwidth}
    \includegraphics[width=1\textwidth,height=\textheight, keepaspectratio]{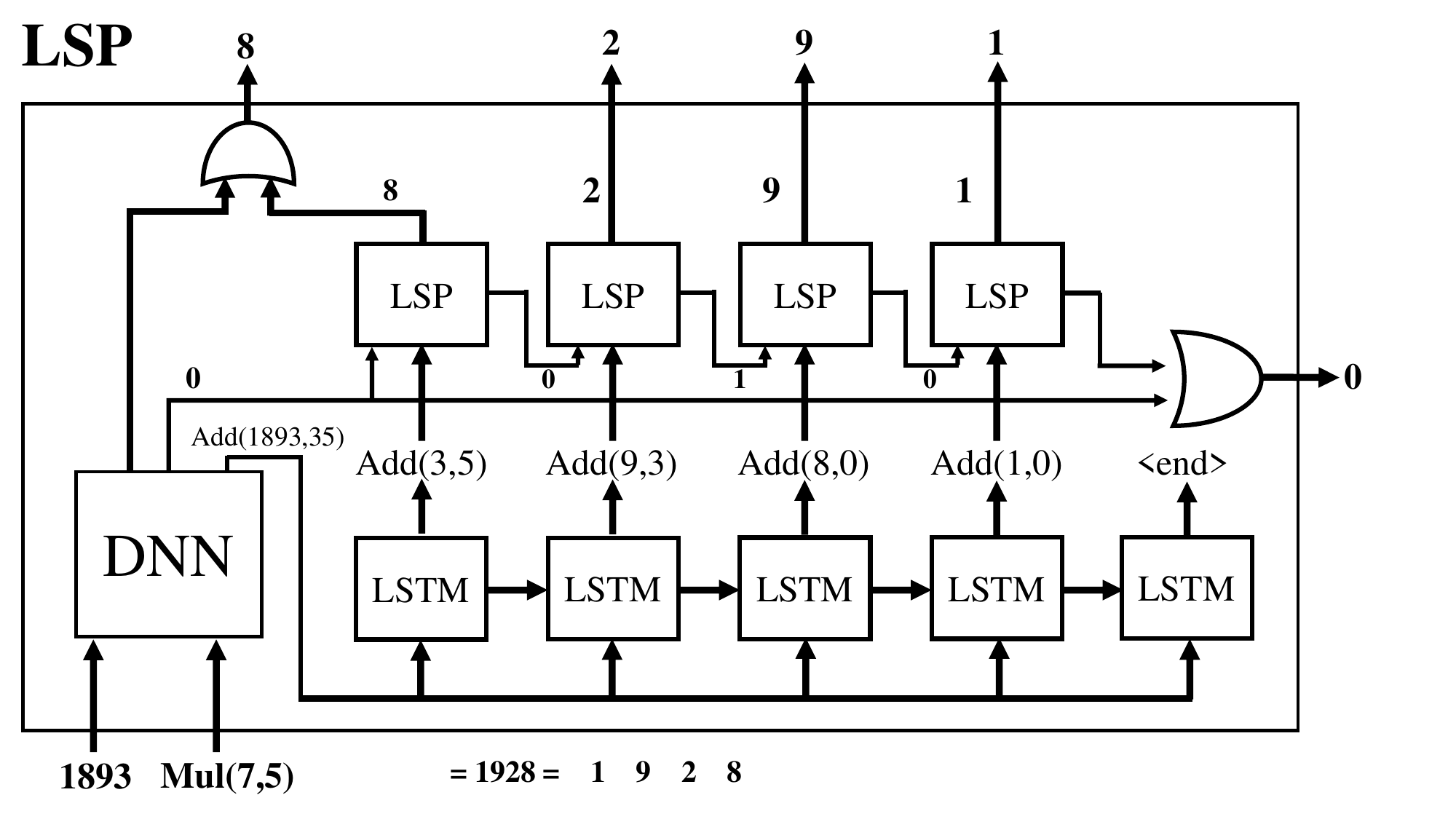}
    \caption{Complex Multiplication 1x1}
    \label{subfig:Complex Multiplication 1x1}
\end{subfigure}
\caption{Examples of complex tasks for multiplication}
\label{fig:Complex Multiplication}
\end{figure}

For an LSP architecture to learn and execute algorithms of more complex problems, it should have basic data structure-processing ability needed for algorithm performance. We propose an LSP architecture with a dynamic memory support (DMS) module to support List and Stack, which are the minimum basic algorithmic data structures.

\begin{figure}[ht!]
\includegraphics[width=0.45\textwidth,height=\textheight, keepaspectratio]{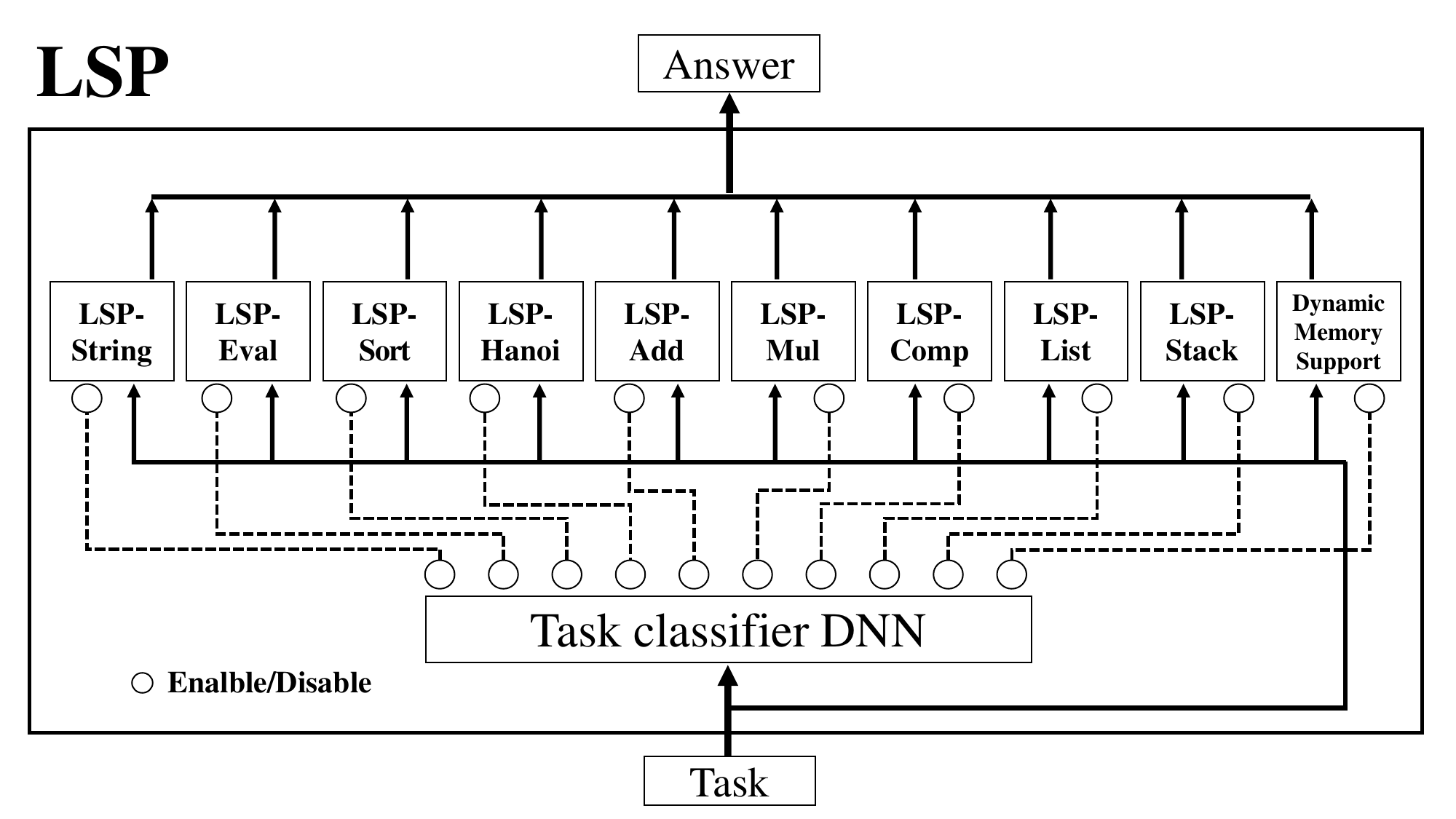}
\caption{(Extended) LSP Architecture}
\label{fig:General LSP Architecture}
\end{figure}

\begin{figure}[ht!]
\includegraphics[width=0.45\textwidth,height=\textheight, keepaspectratio]{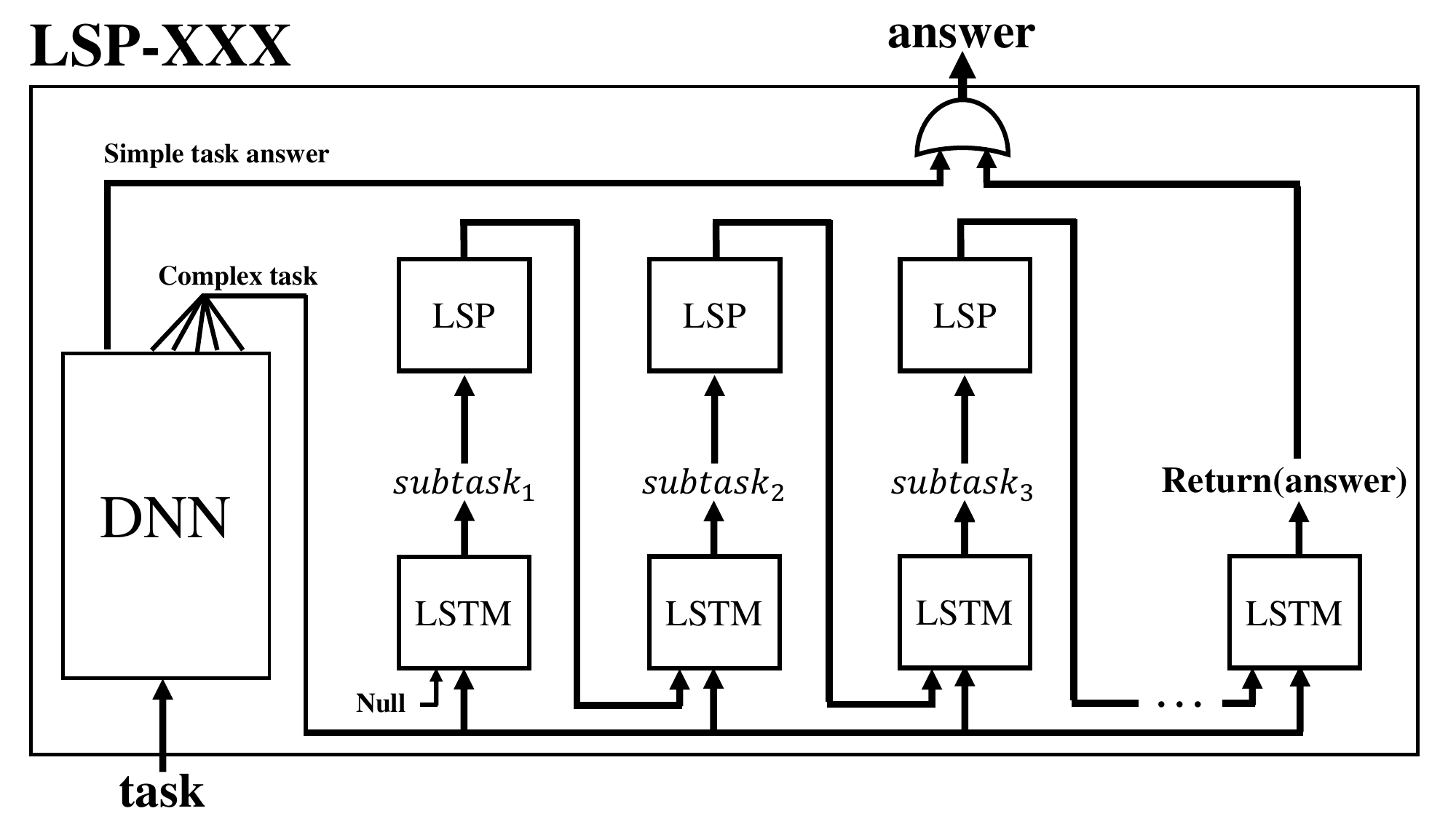}
\caption{LSP-XXX for specific function (XXX is a function name)}
\label{fig:LSP-XXX}
\end{figure}

\noindent LSP consists of many LSP-XXXs that are characterized by their respective functions. LSP distinguishes the input task by function, then transfers it to the corresponding LSP-XXX. The LSP-XXX for each function is a recursive architecture composed of DNN and LSTM. DNN receives input task with input data, and distinguishes whether it is a simple or a complex task. If it is a simple task, DNN outputs the memorized answer. If it is a complex task, DNN sends it to the corresponding LSTM to generate subtasks as a learned solving procedure. Each subtask enters LSP recursively to arrive at an answer. The LSP output answer enters the LSTM input of the next step, thereby controlling the generation and execution of the subtask in the next step.

LSP architecture is highly scalable, because there is no limit to the addition of a new LSP-XXXs, and the task classifier DNN of the LSP automatically finds the corresponding LSP-XXX of each input task.

DMS supports the memory functions required for LSP-List and LSP-Stack, such as read, write, rightShift, leftShift, newList, newStack, isEmpty, etc. LSP-List supports basic list-specific tasks, such as InsertList, Head, and Tail, by using basic functions of DMS. LSP-Stack supports basic stack-specific tasks, such as of Push and Pop.

By using basic data structures (e.g., List and Stack), LSP can successfully learn more complex tasks. To demonstrate, we trained LSP to perform the following complex task.

First, we store numbers in a list to avoid limiting the number of digits. Second, meaningful words are extracted from a string obtained through image analysis (i.e., LSP-String). Third, complex arithmetic expressions consisting of a combination of additions, multiplications, and parentheticals are recognized and computed (i.e., LSP-Eval). Fourth, the listed numbers are sorted (i.e., LSP-Sort). Fifth, we run the Hanoi Tower (i.e., LSP-Hanoi).

The solving procedure for LSP-String, LSP-Eval, LSP-Sort, and LSP-Hanoi tasks are briefly described in the following experiment.

\section{Experiment}
\subsection{Experiment environment\newline}
\noindent All experiments were conducted on Intel Xeon 3.50 GHz CPU with 128 GB DRAM with two NVIDIA GTX 1080, Intel 3.40 GHz CPU with 64 GB DRAM and NVIDIA GTX 1080. The experimental program is implemented using Python and Tensorflow.

\subsection{Experimental goal\newline}
\noindent The purpose of the experiment is proving that LSP can learn solving procedures for complex tasks and correctly execute them. We selected three problems: LSP-Eval, LSP-Sort, and LSP-Hanoi. These tasks are supported by other LSP-XXXs for successful execution. Thus, the LSP consists of a total of nine LSP-XXXs and 31 running tasks.

\subsection{Solving Procedure of Eval task of LSP-Eval\newline}
\noindent For the Eval task, the IF2PF subtask converts the infix expression to a postfix expression, and the CalcPF subtask calculates the postfix expression, outputting the result.

\begin{figure}[!ht]
\includegraphics[width=0.47\textwidth,height=\textheight, keepaspectratio]{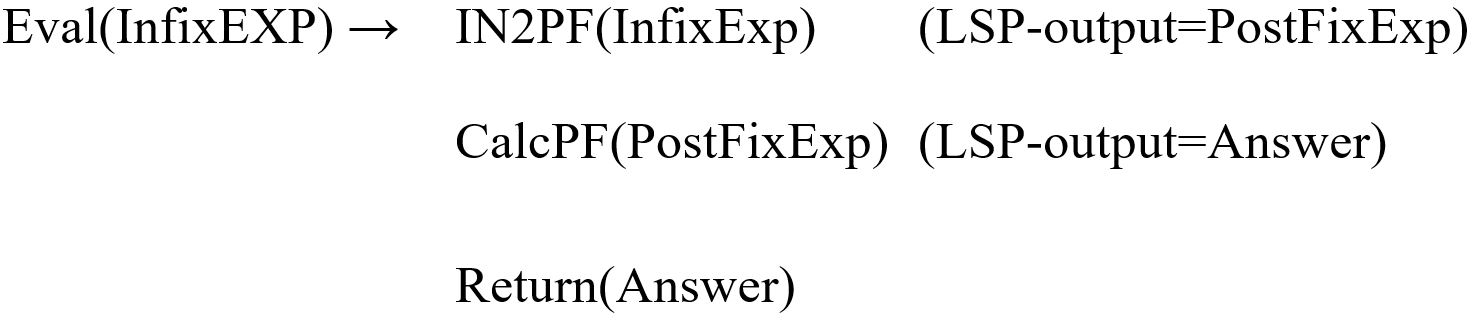}
\end{figure}

\subsection{Solving Procedure of Sort task of LSP-Sort\newline}
\noindent 
The Sort task consists of two solving procedures. First, if the InputList containing the unsorted data is empty, it terminates. Second, if InputList is not empty, it is separated into Head and Tail, sorted by Tail. Then it inserts Head into the ordered Tails to complete sorting. The output of LSP that executes the isEmpty task controls the generation of the next task. That is, an ''if-then-else'' routine is performed.

\begin{figure}[!ht]
\includegraphics[width=0.4\textwidth,height=\textheight, keepaspectratio]{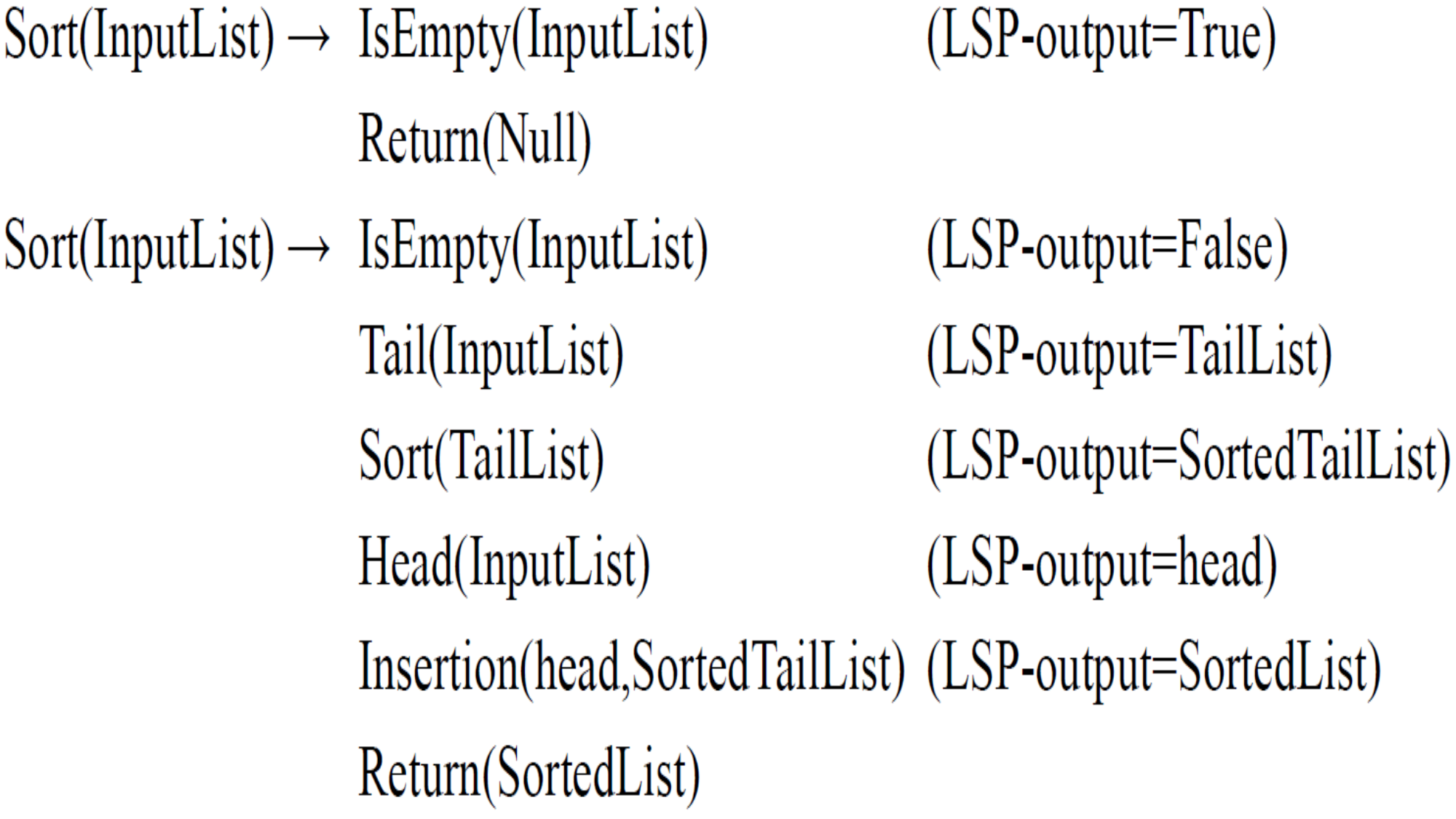}
\end{figure}

\subsection{Solving Procedure of Hanoi task of LSP-Hanoi\newline}
\noindent The Hanoi Task finds the order of moving disks from X queue to Y queue through Z queue, and displays the result of the moves. When moving disks, it is necessary to satisfy the constraint that a large disk cannot be placed atop a small disk. Initially, the DiskList contains the largest disk number in the head and the smaller disks in the tail, in decreasing order. The IsEmpty task controls the creation of the next subtask similar to the sorting task described above. The small disks of Tail are first moved from X to Z, then the largest bottom disk is moved from X to Y. The small disks of Tail are then moved from Z to Y. The Hanoi subtask that moves the disks of Tail is recursively executed by LSP, so that the whole process of moving the disks is performed and output by HanoiOutput subtask.

\begin{figure}[!ht]
\includegraphics[width=0.4\textwidth,height=\textheight, keepaspectratio]{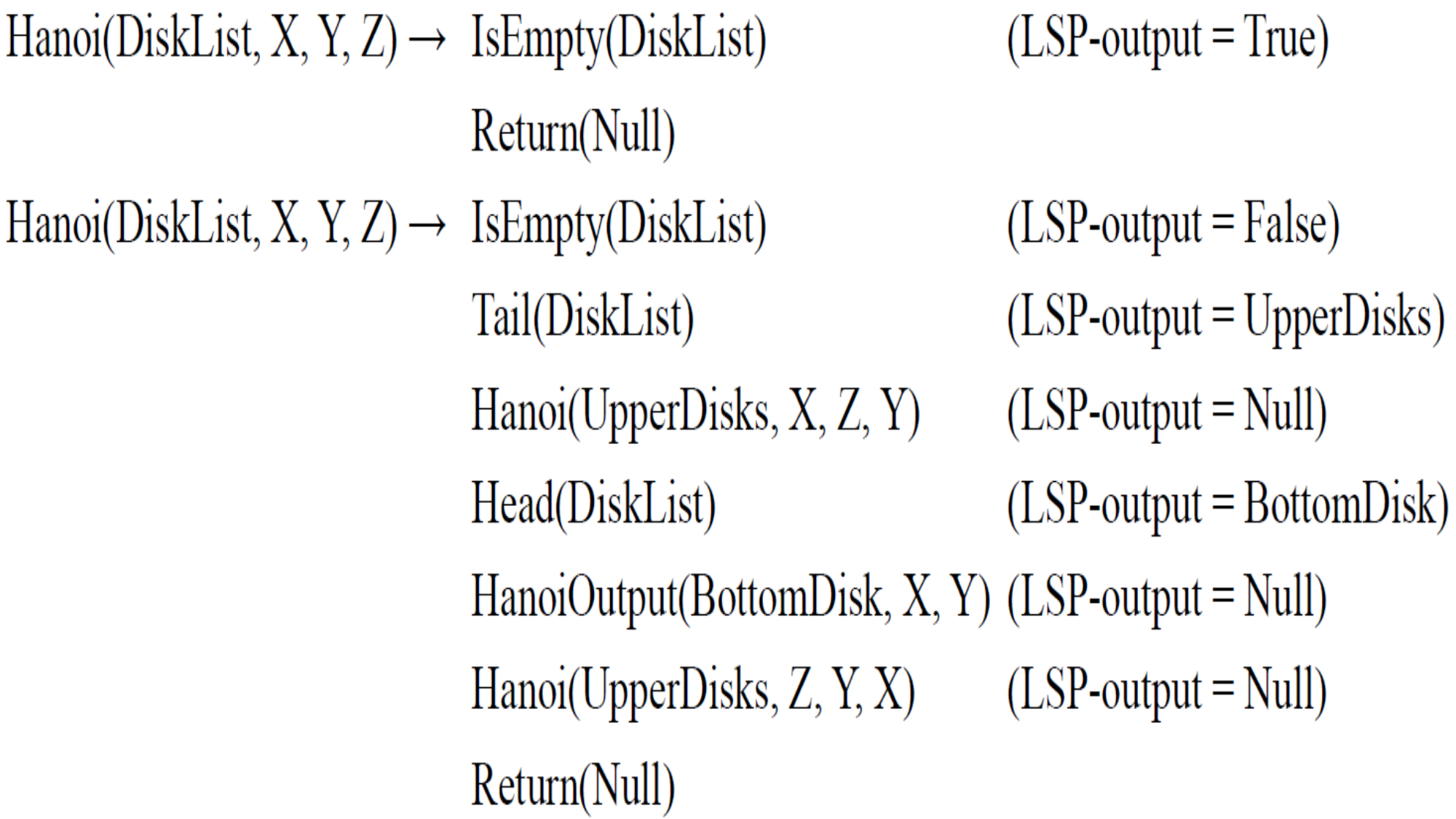}
\end{figure}

\subsection{Training/test data generation of tasks\newline}
\noindent The LSP must have training and test accuracies of nearly 100\% to produce accurate output for the input task. Therefore, the architecture of the DNN and LSTM constituting the LSP is carefully adjusted and the training/test data are carefully selected so that the accuracies are almost 100\% over a total of 31 tasks. DNN uses an average of five layers. The average number of nodes per layer is 164. LSTM uses one-to-two layers, with 96 nodes per layer.

To generate the training/test data for each task, the arguments of the respective subtask(s) and their LSP outputs are regarded as variables, which are separated into independent and dependent types. Because a dependent variable is determined by the value of an independent variable, it does not affect the generation of training data. Thus, training/testing data can be generated by assigning all possible values to all independent variables. We call this method ''all values for all independent variables'' (AVA). However, the size of the training/test data becomes too large. To resolve this, we assign all values for one selected independent variable and randomly assign values to the independent variables. We select all independent variables one-by-one to generate training/test data. We call this method ''all values for one variable, random values for others'' (AVO).

To compare the performance of AVA and AVO, we generated training/test data for a task that has three independent variables, as follows.

There exists {\fontfamily{qcr}\selectfont\small{subtask1(A,B)LSP-Output(C), subtask2(D) LSP-output(E), and return(F)}}. Suppose that $A$, $B$, and $D$ are independent variables; $C$, $E$, and $F$ are dependent variables; and the number of values that can be assigned to each $A$, $B$, and $D$ is 100, respectively. When the training data is generated by the AVA method, the training data totals 1,000,000. However, if the AVO method is used, a total of 30,000 training data are generated. Table 1 shows the experimental results for comparison.

\begin {table}[!ht]
\caption {Comparison of AVA and AVO for the case of 3 variables} \label{tab:Training Data Method Experiment} 
\begin{tabular}{p{0.7cm} p{2.5cm} p{1.0cm} p{1.0cm} p{0.7cm}}\hline
\small{Method} & \small{The Number of Training/Test Data} & \small{Training time} & \small{Training error} & \small{Test error}\\\hline
\small{AVA} & \small{1,000,000/0} & \small{138m 54s} & \small{0\%} & \small{N/A}\\\hline 
\small{AVO} & \small{30,000 /1,000,000} & \small{17m 36s} & \small{0\%} & \small{0\%}\\\hline
\end{tabular}
\end {table}

\begin{figure}[!ht]
\includegraphics[width=0.47\textwidth,height=\textheight, keepaspectratio]{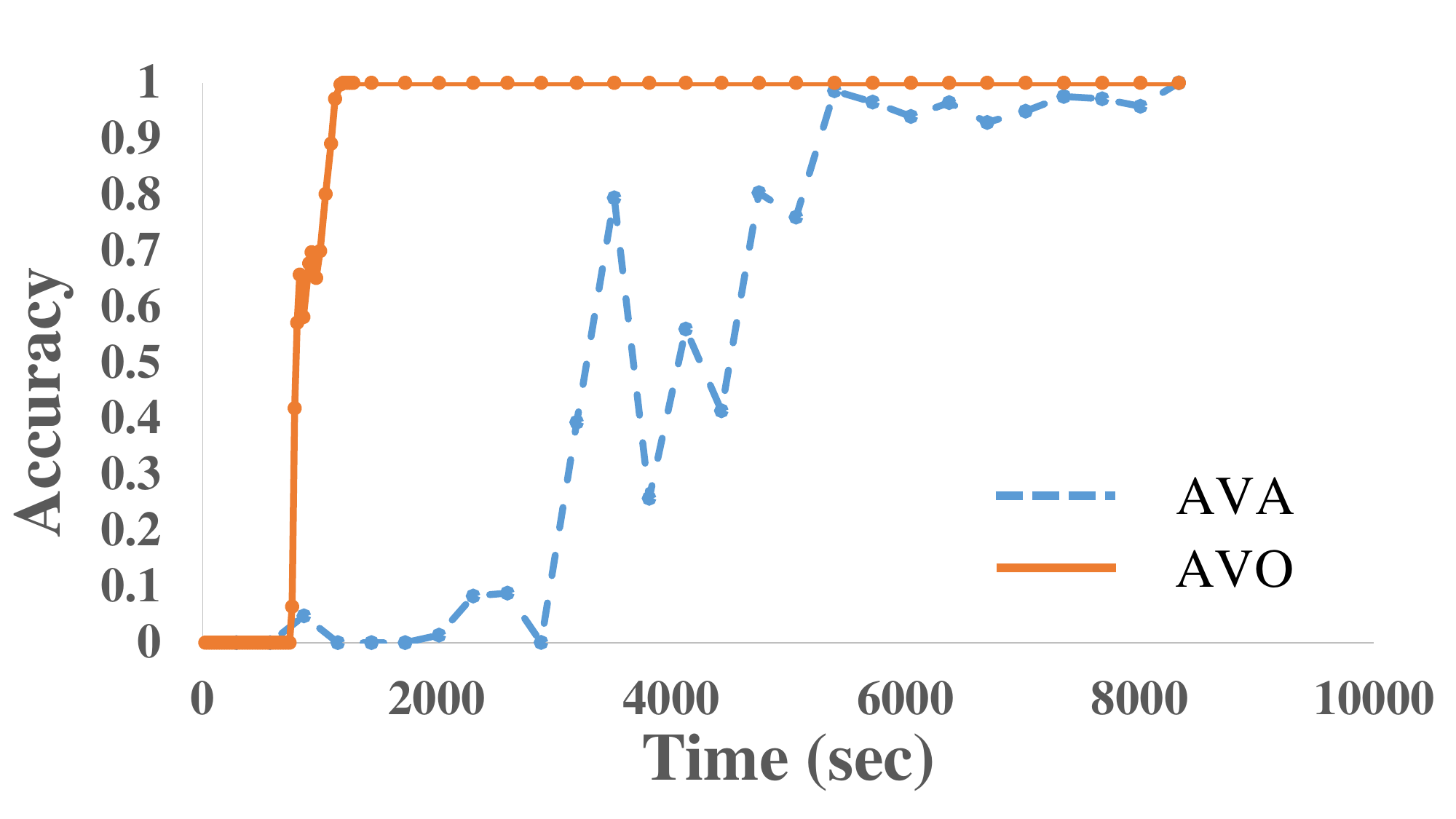}
\caption{Comparison of AVA and AVO}
\end{figure}

The AVO method shows that the training time is much shorter than the AVA method, but the test error is not much worse than AVA.

Training/test data for 31 tasks are generated by the AVO method and are trained by Tensorflow. The average of the training accuracy is 99.9845\%, and the average of the test accuracy is 99.8900\%. Thus, it is trained and tested with almost 100\% accuracy.

\subsection{Integration of Representation learning and Complex reasoning\newline}
This paper demonstrates the utility of ANN complex reasoning by showing that learning algorithms are possible. It is also necessary to show the possibility that representation learning, such as image recognition and complex reasoning, can be naturally integrated. To accomplish this, we input three types of tasks as images, and showed how ANNs solve them. Whereas image input is not a contribution of this paper, it is meaningful as an attempt to show the smooth integration of representation learning and complex reasoning.
We use the modified National Institute of Science dataset: images of handwritten characters in various styles. Using these images, we created 30,000 task images and used them for training. The visual attention-based optical character-reader model is used for image analysis (Shi, et al., 2016). The recognition error rate of the generated task images is close to 0.01\%, and some images that caused recognition errors are excluded. The result of task image analysis is a string. LSP-String converts this string to the task input.

\begin{figure}[!ht]
\centering
\includegraphics[width=0.4\textwidth,height=\textheight, keepaspectratio]{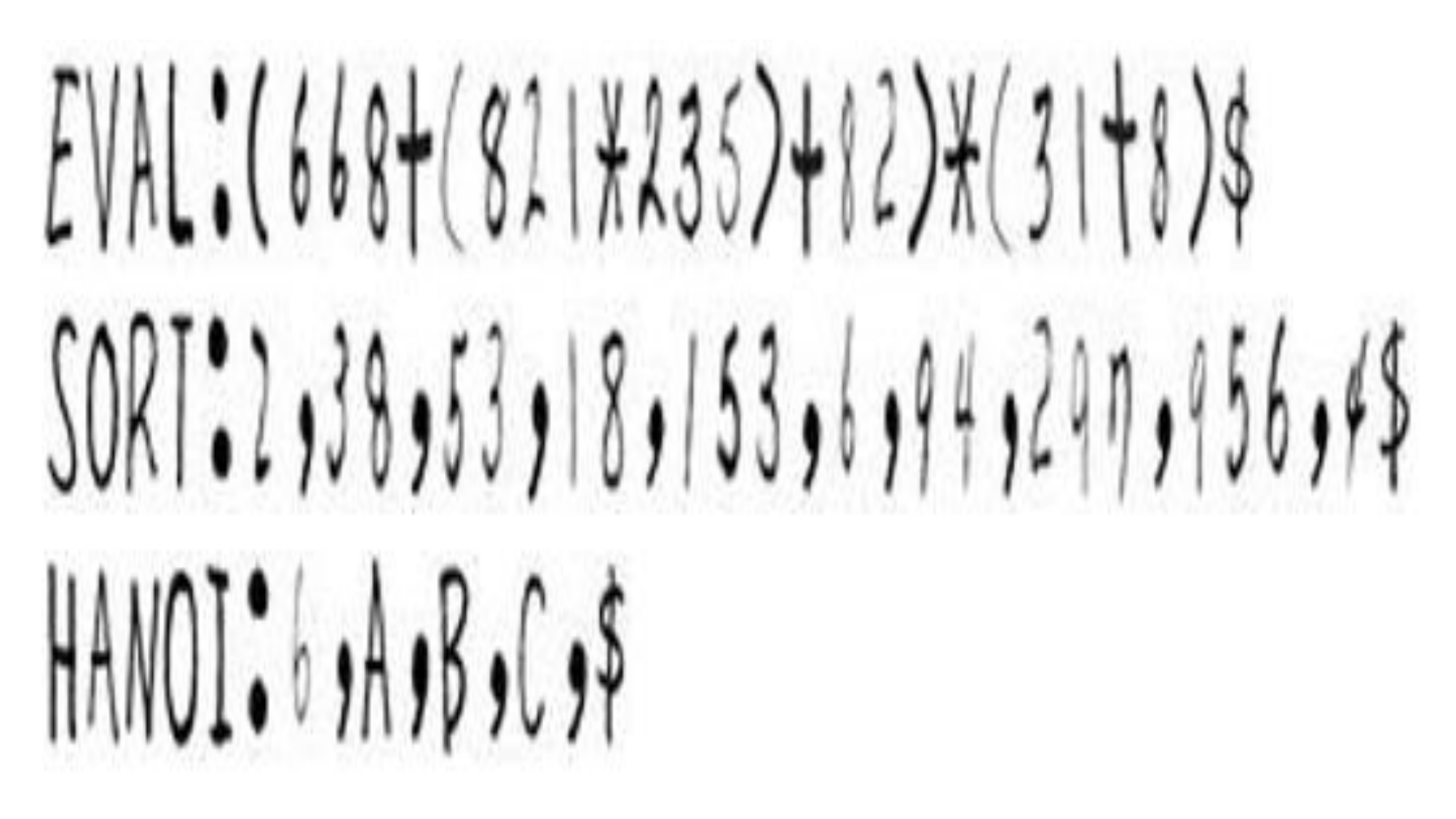}
\caption{Input Images of 3 kinds of tasks }
\end{figure}

As shown in the above example, the performance of the LSP is tested by preparing 200 Eval task images, 100 Sort task images, and eight Hanoi task images using two-to-nine disks.

\begin{figure*}[!ht]
  \centering
  \includegraphics[width=1\textwidth,height=0.7\textheight, keepaspectratio]{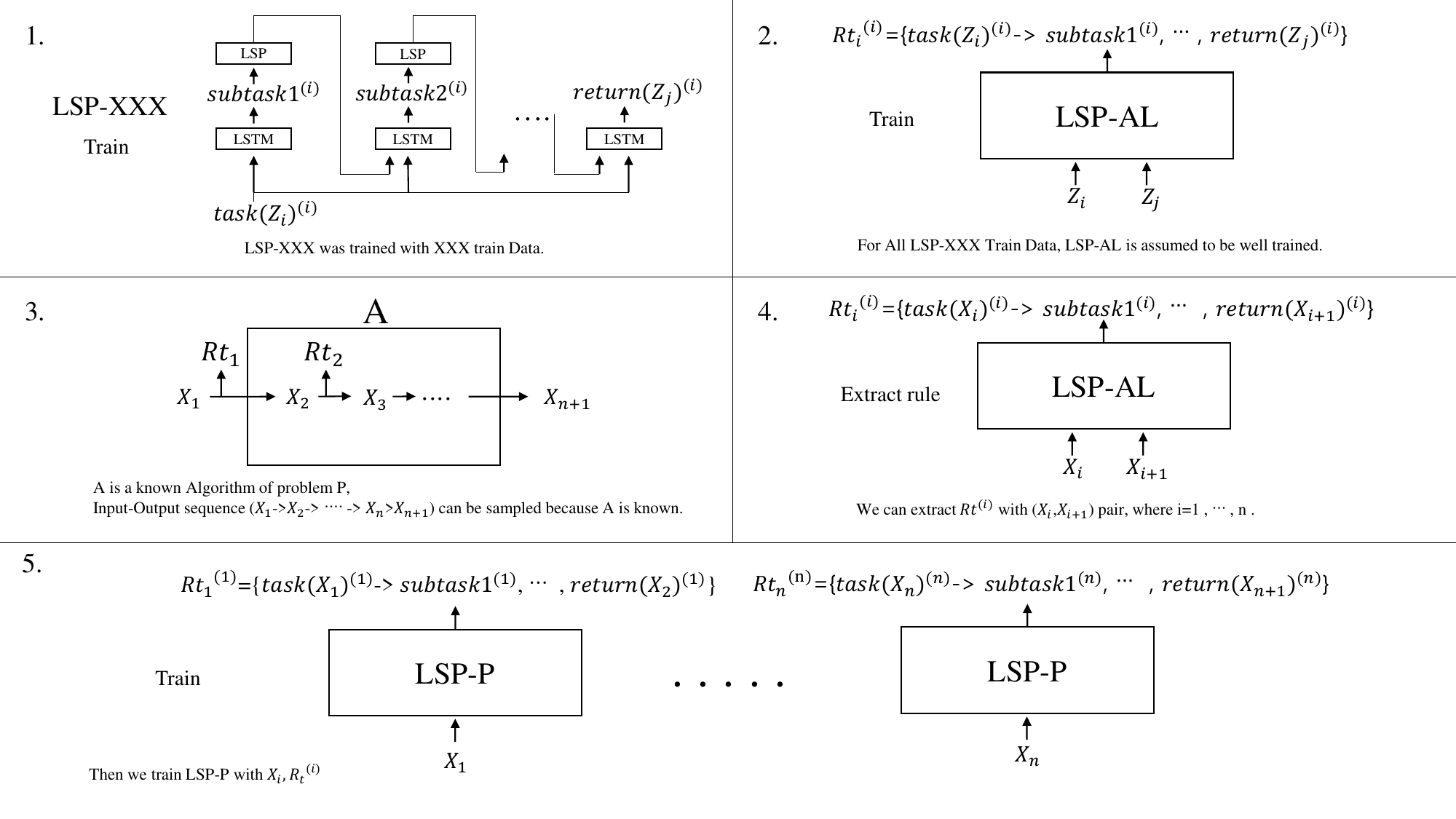}
  \caption{Hypothetical method for algorithm learning LSP for future research.}
  \label{figurelabel}
\end{figure*}

\subsection{Results of Experiment\newline}
\noindent In the Eval task, 191 of 200 output values are correct. Nine cases show a small calculation error, and one digit in the final calculation is incorrect. For example, for the correct output of 27,514,110, an error appears in the hundreds position, as in 27,514,410. There is a possibility that very small errors will accumulate, because the task repeatedly executes multiple subtasks.

We then attempted to sort 10 numbers 100 times. The numbers sorted correctly 98\% of the time, with about a 2\%, error, in which one digit disappears. The cause of the error is estimated to be insufficient coverage of all input cases, owing to lack of training/test data.

For the Hanoi Tower task, we tested eight tasks using two-to-nine disks. All tasks are executed 100\% correctly.

\section{Future Research}

\noindent We now discuss the possibility of future academic progress warranted by LSP, based on the following ideas and hypotheses. We define an algorithm as a sequence of action rules, and we limit our discussion to only those types. We then assume that we can construct algorithms by assembling parts, called ''action rules''.

As shown in this paper, action rules are tasks learned in each LSP-XXX. In this study, we showed that LSP can learn well-known algorithms.

Yet, is there a way to learn the solving procedure of a problem whose algorithm is unknown?

To answer this question, we make the following assumptions. First, LSP contains many LSP-XXXs that are well-trained on their tasks. For each task of every LSP-XXX in the LSP, the training input is the task 's specific input and output pair, and the training output is the task 's action rule (i.e., solving procedure). LSP-AL is trained with the above inputs and outputs.

Assume that the LSP-AL is well-trained on all pre-built LSP-XXXs. Through a well-trained LSP-AL, can we find the solving procedure of the new algorithm (task $A$) in the new problem, $P$, which is not in the existing LSP, as a sequence of tasks in the LSP-XXXs in the LSP?

If we can get the sequence of action rules obtained from specific input and output pairs in task $A$, and we can train the LSP-P for problem, $P$, using them, we can discover the algorithm of problem, $P$.

We can test this hypothesis for a problem, $Q$, whose algorithm is well known, because we already know its solving procedures. If this test is successful for a problem whose algorithm is unknown, but the solving procedures are empirically known for a set of input cases, we conjecture that the LSP proposed in this paper can guess the algorithm for a problem, $P$.

Figure 10 shows a graphical summary for future research.

\section{Conclusion}

\noindent As an attempt to realize complex reasoning, there have been efforts to implement the symbolic reasoning mechanism of von Neumann computers in neural networks via input-output data-learning. Instead, in this paper, we proposed an LSP to learn the solving procedure of a given problem. LSP architecture is a recursive structure consisting of DNN and LSTM, which learns the solving procedure of a given problem and evaluates the answer through a learned procedure. We successfully demonstrated that problem images, such as complex arithmetic expressions, sorting, and Hanoi Tower tasks are recognized by ANN and can be processed with LSP. Thus, we have shown a rudimentary attempt to successfully combine representation learning and complex reasoning. We contend that the human process of problem-solving has been simulated. LSP is different from previous work, because it has advantages, such as flexibility and scalability. We expect future work will learn the solving procedure of each unknown-algorithm problem autonomously, by fully exploiting the scalability of LSP.

\section{References} 

\smallskip \noindent \textit{}Bordes, A.; Chopra, S.; and Weston, J. 2014. Question answering with subgraph embeddings. \textit{Proc. Empirical Methods in Natural Language Processing} arXiv:1406.3676

\smallskip \noindent \textit{}Dahl, G.; et al. 2013. Improving DNNs for LVCSR using rectified linear units and dropout. \textit{ICASSP.}

\smallskip \noindent \textit{}Dayan, P. 2008. Simple substrates for complex cognition. \textit{Frontiers in neuroscience,} 2(2):255.

\smallskip \noindent \textit{}Eliasmith, C. 2013. How to build a brain: A neural architecture for biological cognition. \textit{Oxford University Press.}

\smallskip \noindent \textit{}Farabet, C.; Couprie, C.; Najman, L.; and LeCun, Y. 2013. Learning hierarchical features for scene labeling. \textit{IEEE Trans. Pattern Anal. Mach. Intell.} 35, 1915-1929.

\smallskip \noindent \textit{}Graves, A.; et al. 2016. Hybrid computing using a neural network with dynamic external memory, \textit{nature,} vol 538.

\smallskip \noindent \textit{}Graves, A.; Wayne, G.; and Danihelka, I. 2014. Neural Turing machine. \textit{arXiv}:1410.5401.

\smallskip \noindent \textit{}Hazy, T. E.; Frank, M. J.; and O’Reilly, R. C. 2006. Banishing the homunculus: making working memory work. \textit{Neuroscience,} 139(1):105-118.

\smallskip \noindent \textit{}Jaeger, H. 2016. Deep neural reasoning. \textit{Nature} vol.538 467-468.

\smallskip \noindent \textit{}Hinton, G.; et al. 2012. Deep neural networks for acoustic modeling in speech recognition. \textit{IEEE Signal Processing Magazine} 29, 82–97.

\smallskip \noindent \textit{}Hinton, G. E.; Osindero, S.; Teh, Y. W. 2006. A Fast Learning Algorithm for Deep Belief Nets. \textit{Neural Computation.}18 (7): 1527-1554.

\smallskip \noindent \textit{}Hochreiter, S.; Bengio, Y. ; Frasconi, P.; and Schmidhuber, J. 2001. Gradient flow in recurrent nets: the difficulty of learning long-term dependencies. In Kremer, S. C.; and Kolen, J. F. editors, \textit{A Field Guide to Dynamical Recurrent Neural Networks. IEEE Press}. 

\smallskip \noindent \textit{}Hochreiter, S.; and Schmidhuber, J. 1997. Long short-term memory. \textit{Neural computation,} 9(8):1735-1780.

\smallskip \noindent \textit{}Hochreiter, S. 1991. Untersuchungen zu dynamischen neuronalen Netzen. \textit{Diploma thesis, Institut für Informatik, Technische Univ. Munich.}

\smallskip \noindent \textit{}Hahnloser, R.; Sarpeshkar, R.; Mahowald, M. A.; Douglas, R. J.; Seung, H.S. 2000. \textit{Nature.}405. pp. 947-951.

\smallskip \noindent \textit{}Krizhevsky, A.; Sutskever, I.; and Hinton, G. 2012. ImageNet classification with deep convolutional neural networks. \textit{Proc. Advances in Neural Information Processing Systems} 25 1090-1098.

\smallskip \noindent \textit{}LeCun, Y. 2013. LeNet-5, convolutional neural networks. \textit{http://yann.lecun.com/exdb/lenet/}

\smallskip \noindent \textit{}LeCun, Y.; Bengio, Y.; and Hinton, G. 2015. Deep Learning. \textit{Nature} vol.521 436-444.

\smallskip \noindent \textit{}Mikolov, T.; Deoras, A.; Povey, D.; Burget, L.; and Cernocky, J. 2011. Strategies for training large scale neural network language models. \textit{Proc. Automatic Speech Recognition and Understanding} 196-201.

\smallskip \noindent \textit{}Sainath, T.; Mohamed, A.-R.; Kingsbury, B.; and Ramabhadran, B. 2013. Deep convolutional neural networks for LVCSR. \textit{Proc. Acoustics, Speech and Signal Processing} 8614-8618.

\smallskip \noindent \textit{}Shi, B.; Wang, X.; Lyu, P.; Yao, C.; Bai, X. 2016. Robust Scene Text Recognition with Automatic Rectification.  \textit{CVPR.}

\smallskip \noindent \textit{}Srivastava, N.; Hinton, G.; Krizhevsky, A.; Sutskever, I.; Salakhutdinov, R. 2014. Dropout: A Simple Way to Prevent Neural Networks from Overfitting. \textit{Journal of Machine Learning Research} 15 1929-1958.

\smallskip \noindent \textit{}Szegedy, C.; et al. 2014. Going deeper with convolutions. \textit{The IEEE Conference on Computer Vision and Pattern Recognition (CVPR)}, 2015, pp. 1-9

\smallskip \noindent \textit{}Reed, S.; and de Freitas, N. 2015. Neural Programmer-Interpreters. \textit{arXiv}:1511.06279.

\smallskip \noindent \textit{}Tetko, I. V.; Livingstone, D. J.; Luik, A. I. 1995. Neural network studies. 1. Comparison of Overfitting and Overtraining. \textit{J. Chem. Inf. Comput. Sci.} 35 (5): 826-833. 

\smallskip \noindent \textit{}Tompson, J.; Jain, A.; LeCun, Y.; and Bregler, C. 2014. Joint training of a convolutional network and a graphical model for human pose estimation. \textit{Proc. Advances in Neural Information Processing Systems} 27 1799-1807.

\smallskip \noindent \textit{}TWerbos, P. 1974. Beyond Regression: New Tools for Prediction and Analysis in the Behavioral Sciences. \textit{PhD thesis, Harvard University.}



\bibliographystyle{siam}
\bibliography{ref-short}

\end{document}